\documentclass[conference,a4paper]{IEEEtran}
\IEEEoverridecommandlockouts
\usepackage{CJKutf8}
\usepackage{cite}
\usepackage{hyperref}
\usepackage{subfigure}
\usepackage{algorithm}
\usepackage{algpseudocode}
\usepackage{threeparttable}
\usepackage{amsmath,amssymb,amsfonts}
\usepackage{graphicx}
\usepackage{textcomp}
\usepackage{xcolor}
\def\BibTeX{{\rm B\kern-.05em{\sc i\kern-.025em b}\kern-.08em
    T\kern-.1667em\lower.7ex\hbox{E}\kern-.125emX}}
\begin{document}

\title{A Gis Aided Approach for Geolocalizing an Unmanned Aerial System Using Deep Learning}

\author{\IEEEauthorblockN{Jianli Wei, Deniz Karakay, Alper Yilmaz, \textit{Senior Member, IEEE}}
\IEEEauthorblockA{\textit{Photogrammetric Computer Vision Lab., The Ohio State University,} Columbus, OH, USA \\
\textsuperscript{}\big\{wei.909, yilmaz.15\big\}@osu.edu, deniz.karakay@metu.edu.tr}}

\maketitle

\begin{abstract}
The Global Positioning System (GPS) has become a part of our daily life  with the primary goal of providing geopositioning service. For an unmanned aerial system (UAS), geolocalization ability is an extremely important necessity which is achieved using Inertial Navigation System (INS) with the GPS at its heart. Without geopositioning service, UAS is unable to fly to its destination or come back home. Unfortunately, GPS signals can be jammed and suffer from a multipath problem in urban canyons. Our goal is to propose an alternative approach to geolocalize a UAS when GPS signal is degraded or denied. Considering UAS has a downward-looking camera on its platform that can acquire real-time images as the platform flies, we apply modern deep learning techniques to achieve geolocalization. In particular, we perform image matching to establish latent feature conjugates between UAS acquired imagery and satellite orthophotos. A typical application of feature matching suffers from high-rise buildings and new constructions in the field that introduce uncertainties into homography estimation, hence results in poor geolocalization performance. Instead, we extract GIS information from OpenStreetMap (OSM)\cite{OpenStreetMap} to semantically segment matched features into building and terrain classes. The GIS mask works as a filter in selecting semantically matched features that enhance coplanarity conditions and the UAS geolocalization accuracy. Once the paper is published our code will be publicly available at \url{https://github.com/OSUPCVLab/UbihereDrone2021}.
\end{abstract}


\section{Introduction}
\label{intro}
UAS geolocalization in a GPS-denied environment is a very hard problem especially when a monocular camera is the only sensor onboard. Often times, it is insufficient to achieve UAS geolocalization as the single camera often yields sub-optimal solution leading to trajectory drifts and scale inconsistency ~\cite{ELHASHASH202262, zhang2018scale}. To overcome these drawbacks, a widespread and general solution is to add additional sensors such as stereo camera, inertial measurement units or LiDAR equipment. The typical approach to using cameras is to extend 2D RGB image to 2.5D RGBD with depth information. This approach, however, has several drawbacks. First, the depth estimation introduces additional propagating errors, which result in drifts, hence, bad performance and accuracy. Second, most existing approaches require very high video frame rate (FPS) rate than the standard 24 fps. For instance, the SVO ~\cite{forster2014svo} requires FPS larger than 60. Higher FPS usually promises better performance especially when UAS rapidly moves relative to its flying height ~\cite{mur2015orb, mur2017orb, campos2021orb}. Third, additional sensor requirements mean accurate lever-arm and boresight estimation and higher cost. Considering those drawbacks, we propose a new approach to UAS geolocalization with three main contributions. First, it eliminates the need to use a series of consecutive frames. The UAS geoposition at time \textit{t} is estimated only using the image at time \textit{t}. There will not be drift concern anymore. Second, due to the first contribution, our approach is able to geolocalize UAS from video or taken images stream. Higher speed still requires higher FPS. However, 30 FPS, as all recording camera supports, is high enough for our experiment. Third, our approach does not require a stereo camera rig. The only requirement is that the camera is face-down direction.

The rest of this paper is organized as follows. Section \ref{sec:related} reviews related work in image matching and GIS. Section \ref{sec:problem} introduces the problem and discusses the details of the proposed approach. Section \ref{sec:experiment} demonstrates the performance and Section \ref{conc} concludes the paper.

\section{Related Work}
\label{sec:related}

Image matching is an active area of research in computer vision which includes three main phases: keypoint detection, descriptor assignment and matching. Keypoint descriptors are generally categorized into handcrafted descriptors and deep learning descriptors~\cite{ma2021image}. Handcrafted descriptors highly rely on expert knowledge to model local texture variation. So far, there are still many traditional handcrafted descriptors that are widely used in many applications. SIFT and SURF are the two most commonly used techniques ~\cite{lowe2004distinctive, bay2008speeded}. SIFT adopts difference of Gaussian (DoG) kernel and use local maxima in space and scale to achieve scale invariance. DoG computation is very time-consuming due to float kernel variance. SURF is therefore proposed to approximate gradient computation by using Harr wavelet decomposition in $x$ and $y$ directions. Integral images replacing DoG float kernel variance accelerate computation over SIFT. Apart from handcrafted, deep descriptors are recently proposed as a self-supervised learning problem using modern deep learning techniques to learn a representation vector similar to handcrafted descriptors. Kumar \textit{et al.} ~\cite{kumar2016learning} implemented a triplet loss and Siamese network to increase positive and negative pair margins. DeTone \textit{et al.} proposed SuperPoint as a self-supervised framework for learning point of interest especially for cross-domain descriptors such as synthetic-to-real. Cross-domain matching ability is necessary to our problem as it could mitigate the dissimilarity between Google satellite images and UAS-taken images that contain illumination variation and season differences. SuperGlue ~\cite{sarlin2020superglue} as an end-to-end framework that integrates SuperPoint as feature descriptor and a graph neural network to find correspondences and reject unmatched keypoints among two local images.

To estimate the motion of the platform, homography is generally used which estimates the transformation of planar surfaces viewed across two camera viewpoints. In UAS geolocalization, use of homography or other plane transformations requires the matched keypoints to be coplanar in 3D. From a down-looking camera installed on UAS, most keypoints lie on objects, e.g. road, trees, pedestrians and vehicles, are close to the ground plane, hence they are quasi-coplanar. However, the buildings in the images vary in height and are higher than ground place breaking the coplanarity requirement. Therefore, we seek to exclude matching keypoints that belong to buildings. For this purpose, we exclusively use the Geographic Information Systems (GIS), in particular, the Open Street Map (OSM)\cite{OpenStreetMap}. OSM as a free wiki world map provides different geotagged GIS layers, such as the building footprints in georeferenced satellite images, such as the Google orthophotos generated from satellite images.

\section{Methodology}
\label{sec:problem}
Geolocating a UAS based on images is a challenging task that requires the selection of a georeference frame as basemap, defining a geoposition reference point (similar to GPS); acquiring map-based semantic mask with building and ground separated into two classes, choosing matching keypoints to estimate the homography matrix, and transforming pixel position to GPS.

\begin{figure*}[htbp!]
\centerline{\includegraphics[width=1.82\columnwidth]{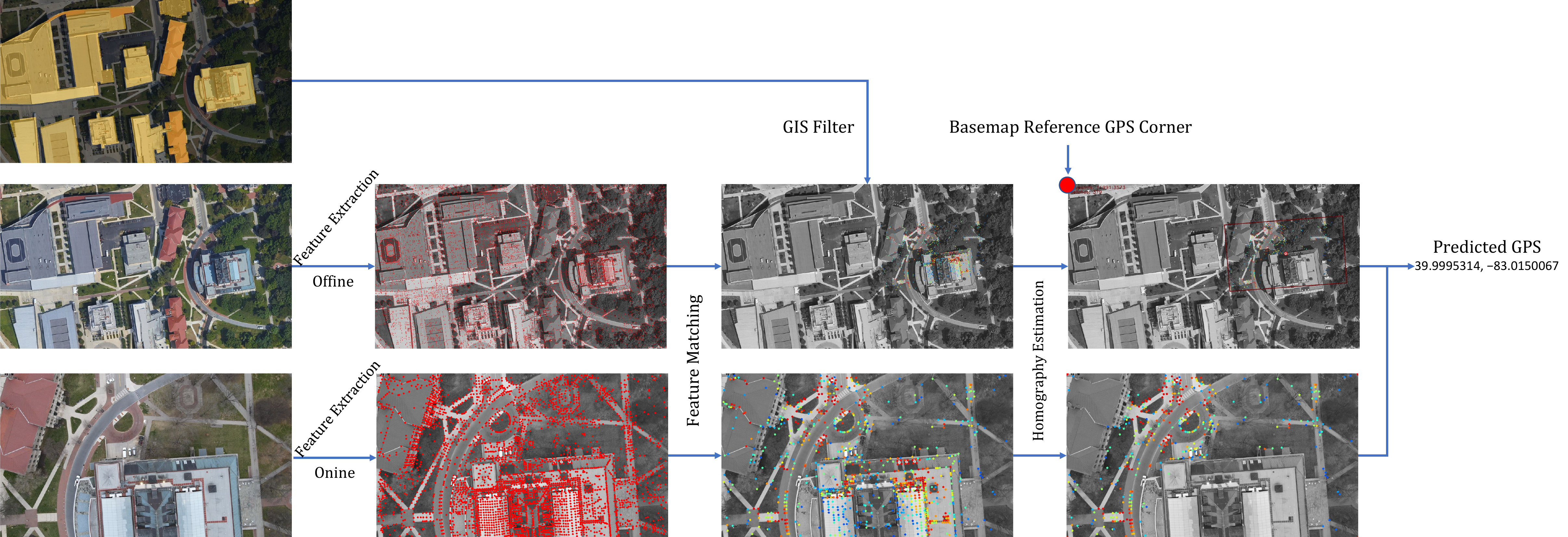}}
\caption{Flow diagram showing the relation between the image matching, GIS and geopositioning modules.}
\label{Flowchart}
\end{figure*}

To address these challenges, we propose an end-to-end framework that takes UAS images and satellite image-based basemap as inputs and generates geolocation in GPS coordinates as output. The framework includes an image matching module, GIS module and geopositioning module shown as illustrated in Fig. \ref{Flowchart}. As for designing image matching module, we utilize pretrained SuperGlue network as our backbone network to extract features respectively from UAS taken image and satellite basemap and match them.

\subsection{Image matching}
\label{ImgMatching}
The proposed algorithm starts with finding a reference basemap with which the UAS acquired image has a geometric relation. In our implementation we used orthophotos from Google Earth as the satellite map due to its contextual similarity to a UAS mounted camera looking down direction. This similarity provides overlapping map layers such as vegetation, road network and buildings. The API to Google Earth handily provides access through a Python script.

Considering basemap and UAS taken images are contextually similar, latent relationship between them could be established efficiently through use of common keypoint extractors. The descriptor used by these extractors, however, cannot undertake the contextual similarity task due to the fact that they are too general and are susceptible to illumination variation, seasonal changes and spatiotemporal changes from construction. To achieve better keypoint descriptors and matching, we utilize a more recent matching algorithm that extracts resilient keypoints and descriptors and provides better matching accuracy (see positive matching keypoint pairs in Fig \ref{Flowchart}). The algorithm we adopt is the SuperGlue ~\cite{sarlin2020superglue} which includes SuperPoint and a pretrained GNN as keypoint detector and matching framework respectively. 

\subsection{Integration with GIS}
\label{GIS}

The matching keypoints generally lie in various object categories including the ground plane and road network. The superglue algorithm is ignorant of the keypoint categories. Due to geometric reasons, the keypoints on buildings break the required planarity condition for the homography estimation. To mitigate this shortcoming, we label the matched keypoints into building and ground plane categories to ensure coplanarity condition during the estimation process. The proposed approach proceeds by integrating the GIS footprint from the OSM as shown in Fig \ref{Flowchart}. The marked yellow area covers the building layer without any information on their heights, and the dark area depicts the ground plane. GIS integration module filters out matching keypoints laying on the buildings. In the case when the UAS flies over sites that only contain matching keypoints that are on the buildings, the proposed approach considers an iterative scheme to select keypoint matches of buildings with similar heights (see Algo. \ref{alg:cap} for the selection of valid matching keypoints for homography estimation).

\begin{algorithm}
\caption{GIS integration}\label{alg:cap}
\textbf{Input: $f_{Ground},f_{Building}$} \\
\textbf{Output: $f_{Valid}$}
\begin{algorithmic}
\If{$N_{Ground}\geq T $}
    \If {$N_{Building}/N_{Ground}<3 $}
        \State $f_{Valid}=f_{Ground}$
    \Else
        \State $f_{Valid}=f_{Building}$
    \EndIf
\ElsIf{$N_{Buidling}\geq T $}
    \State $f_{Valid}=f_{Building}$
\Else
    \State $f_{Valid}=f_{Ground}\cup{f_{Building}}$
\EndIf
\end{algorithmic}
\end{algorithm}

where \textit{$f_{Ground}$} and \textit{$f_{Building}$} are matched features belonging to ground and buildings, \textit{$f_{Valid}$} are valid matched features after Gis filter, \textit{$N_{Ground}$} and \textit{$N_{Building}$} represent number of those features respectively. \textit{T} as threshold equals to 50 experimentally.

\subsection{Geopositioning}
\label{Geo}
After the GIS integration phase, the validated matching keypoints are in the local UAS image coordinates and global basemap coordinates. For simplicity, we denote them as $P_{Image}$ and $P_{Basemap}$. The homography matrix is then estimated as:
\begin{equation}
    P_{Basemap} = h_{3\times3}*P_{Image}
\end{equation}
We use a random sample consensus approach that uses four matching keypoints at a time to project the UAS taken image to the basemap coordinate system as a quadrilateral, whose moment of inertia is defined as UAS's central position in pixel in reference to basemap left-top corner which will then be transformed to GPS coordinate.

\section{Experiments}
\label{sec:experiment}

Without additional training, we deployed the pre-trained SuperGlue model into our problem. We observed that even under different viewing angles and illumination conditions, SuperGlue achieved an excellent image matching performance. In the next discussion, we will explain the collected dataset and the geopositioning results.

\subsection{Dataset}
\label{subsec: datasets}
Since there is no existing UAS geolocalization benchmark, we collected UAS data by flying a DJI Mavic Air2 at our university campus. The images are taken with a speed of  2.5 meters/second, corresponding to approximately 3-meter UAS motion between two images. Ground-truth (GT) geolocation is obtained from the GPS sensor on the drone with reported accuracy of $\pm2.5$ meters. The UAS is set to be oriented in the north direction for image acquisition during the flight. The basemap keypoint extraction is performed offline to reduce inference time.

\subsection{Results}
\label{subsec: experiment results}

\begin{figure}[htbp]
\centering    
\subfigure[Predicted trajectory without GIS integration (blue) and GT (red)]{
\label{fig:2a}
\includegraphics[width=0.88\columnwidth]{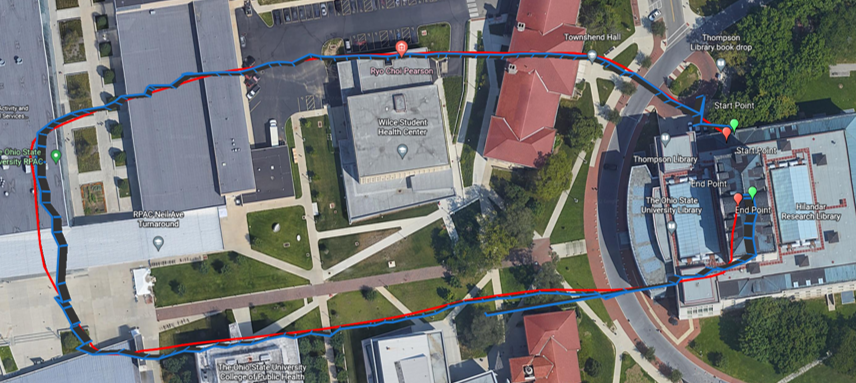}}
\subfigure[Predicted trajectory with GIS integration (yellow) and GT (red)]{
\label{fig:2b}
\includegraphics[width=0.88\columnwidth]{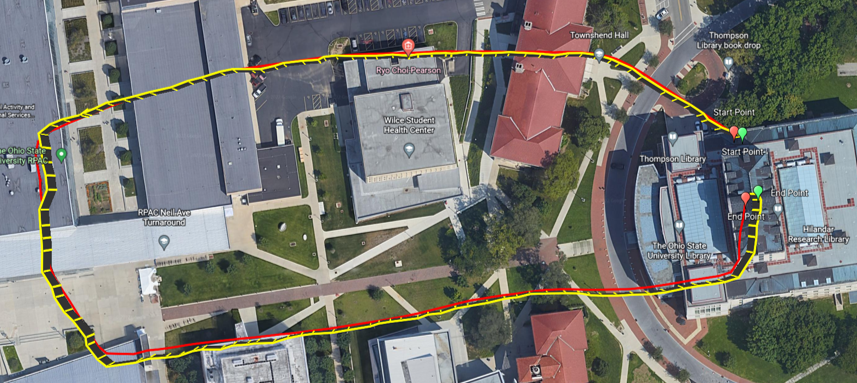}
}
\caption{UAS predicted trajectory and the ground truth superimposed on the Google Earth for visualization.}     
\label{result1}     
\end{figure}

In Figure \ref{result1}, we show the predicted UAS trajectory without GIS integration \ref{fig:2a} and with GIS integration against the ground truth GPS position acquired from UAS acquired images \ref{fig:2b}. It can be observed the yellow trajectory generated from the proposed approach is smoother and fits better with the red ground truth trajectory compared with the blue one in\ref{fig:2a}. Quantitatively, we evaluate UAS geolocalization accuracy of the predicted geolocation against ground truth GPS in Table \ref{tab1}. Overall, GIS integration reduces the error by an average of about 31.88\%. Our proposed approach could achieve UAS geopositioning within an average error of 3.96 meters which is within the error margin of the ground-truth data.

\begin{table}[htbp]
\centering
\caption{UAS geopositioning quantitative analysis in mean absolute error (MAE) and maximum (Max).}
\begin{threeparttable}
\begin{tabular}{cccc}
\hline
Seq. & MAE/Max(m) w/o. GIS & MAE/Max(m) w/. GIS & MAE Imp. \\
\hline
1 & 4.99 / 76.41 & 3.32 / 10.00 & 33.47\% \\
2 & 6.09 / 25.50 & 4.49 / 11.55 & 26.27\% \\
3 & 6.35 / 36.09 & 4.07 / 24.32 & 35.90\% \\
\hline
\end{tabular}
\end{threeparttable}
\label{tab1}
\end{table}

\section{Conclusions and Future Work}
\label{conc}
Our approach achieves UAS geopositioning using embedded monocular camera in an end-to-end manner and a mean absolute error within 4 meters. Considering UAS itself GPS ground truth accuracy range of $\pm2.5$ meters, the resulting trajectory is acceptable. We noticed that setting UAS to acquire images in the north direction may limit the applicability in real-world applications, which we will tackle as the next step.

\bibliographystyle{ieeetr}
\bibliography{Main}

\end{document}